\documentclass{article}
\PassOptionsToPackage{numbers}{natbib}

\usepackage[preprint]{neurips_2026}

\usepackage[utf8]{inputenc} 
\usepackage[T1]{fontenc}    
\usepackage{hyperref}       
\usepackage{url}            
\usepackage{booktabs}       
\usepackage{amsfonts}       
\usepackage{nicefrac}       
\usepackage{microtype}      
\usepackage{xcolor}         

\usepackage{graphicx}
\usepackage{tabularx}
\usepackage{array}
\usepackage{amsmath}
\usepackage{amssymb}
\usepackage{tcolorbox}
\usepackage{enumitem}
\newtheorem{definition}{Definition}
\newtheorem{proposition}{Proposition}
\newtheorem{claim}{Claim}
\title{Agentic AIs Are the Missing Paradigm \\ for Out-of-Distribution Generalization \\ in Foundation Models}

\author{%
  Xin Wang\thanks{Corresponding author.} \\
  Tsinghua University \\
  \texttt{xin\_wang@tsinghua.edu.cn} \\
  \And
  Haibo Chen \\
  Tsinghua University \\
  \texttt{chb24@mails.tsinghua.edu.cn} \\
  \And
  Wenxuan Liu \\
  Tsinghua University \\
  \texttt{liuwx20@mails.tsinghua.edu.cn} \\
  \And
  Wenwu Zhu\footnotemark[1] \\
  Tsinghua University \\
  \texttt{wwzhu@tsinghua.edu.cn} \\
}

\begin{document}

\maketitle

\begin{abstract}
Foundation models (FMs) are increasingly deployed in open-world settings where distribution shift is the rule rather than the exception. The out-of-distribution (OOD) phenomena they face---knowledge boundaries, capability ceilings, compositional shifts, and open-ended task variation---differ \emph{in kind} from the settings that have shaped prior OOD research, and are further complicated because the pretraining and post-training distributions of modern FMs are often only partially observed. \textbf{Our position is that OOD for foundation models is a structurally distinct problem that cannot be solved within the prevailing model-centric paradigm, and that agentic systems constitute the missing paradigm required to address it.} We defend this claim through four steps. First, we give a stage-aware formalization of OOD that accommodates partially observed multi-stage training distributions. Second, we prove a \emph{parameter coverage ceiling}: there exist practically relevant inputs that no model-centric method (training-time or test-time) can handle within tolerance $\varepsilon$, for reasons intrinsic to parameter-based representation. Third, we characterize agentic OOD systems by four structural properties---perception, strategy selection, external action, and closed-loop verification---and show that they strictly extend the reachable set beyond the ceiling. Fourth, we respond to seven counterarguments, conceding two, and outline a research agenda. We do not claim that agentic methods subsume model-centric ones; we argue that the two are complementary, and that progress on FM-OOD requires explicit recognition of the agentic paradigm as a first-class research direction.
\end{abstract}

\section{Introduction}
\label{sec:intro}

Out-of-distribution (OOD) generalization has been a central concern in machine learning for more than a decade~\cite{learning2009dataset,liu2021towards}. Once models leave curated benchmarks for open-world deployment, distribution shift becomes both routine and consequential: medical AI systems encounter patient populations whose demographics and imaging hardware differ from the training cohort~\cite{finlayson2021clinician}; autonomous vehicles meet weather and traffic configurations no finite training set enumerates~\cite{filos2020can}; language models are queried about topics, events, and formats absent from pretraining~\cite{kandpal2023large}.

Foundation models (FMs)~\cite{bommasani2021opportunities}---including language~\cite{brown2020language,touvron2023llama}, vision~\cite{radford2021learning,kirillov2023segment}, and graph models~\cite{liu2023towards,mao2024position}---face OOD phenomena that go \emph{qualitatively} beyond classical assumptions. First, training is multi-stage: pretraining, supervised fine-tuning, and post-training each induce distinct distributions~\cite{ouyang2022training}, so a deployment input may be ID for one stage yet OOD for another. Second, many FM pretraining corpora are undisclosed or partially documented, so the classical comparison between a known source $P$ and test $Q$ must be reformulated when $P$ itself is only partially observed.

Traditional benchmarks~\cite{koh2021wilds,hendrycks2019benchmarking} assume a fixed task with bounded covariate, label, or domain shift. FM deployment breaks these premises: tasks are open-ended, shifts compose, and generative failures are often \emph{silent}, producing fluent wrong outputs rather than abstentions~\cite{ji2023survey,huang2025survey}.

Existing OOD methods, though diverse in mechanism, share a common structural commitment. Training-time methods---invariant learning~\cite{arjovsky2019invariant,krueger2021out}, distributionally robust optimization~\cite{sagawa2019distributionally}, and data augmentation~\cite{hendrycks2019augmix}---shape the model before deployment. Test-time methods---test-time adaptation (TTA)~\cite{wang2020tent,liang2020we}, test-time training (TTT)~\cite{sun2020test}, and domain adaptation~\cite{ganin2016domain}---adjust the model at deployment using target-side signals (unlabeled inputs, pseudo-labels, auxiliary objectives). Whichever variable they tune, they all share one assumption: \emph{when a distribution shift occurs, the right response is to adjust the model}. We call this commitment \textbf{model-centrism}.

Model-centrism breaks down in three cases formalized below. (i)~\emph{Knowledge gaps}: target-side signals alone cannot inject information absent from all relevant $\mathcal{D}^{(k)}$. (ii)~\emph{Capability gaps}: some computations are better delegated to tools than encoded in practically reachable parameters, e.g., multi-digit arithmetic~\cite{dziri2023faith}. (iii)~\emph{Compositional shifts}: a medical or legal query may combine an unfamiliar subgroup, a recent guideline, and an unseen task format, while single-shift methods degrade when shifts coincide~\cite{wiles2022fine}. Some OOD failures require consulting a reference, using an instrument, decomposing the problem, or acknowledging uncertainty.

\paragraph{Position (informal).} \emph{OOD for foundation models is a structurally distinct problem that the model-centric paradigm cannot fully solve. Agentic systems---defined by perception of their own distributional state, selection among heterogeneous response strategies, invocation of external resources, and closed-loop verification---constitute the missing paradigm. The two paradigms overlap on inference-time model adjustment, but each contains actions outside the other's reach; they are complementary, not competing.}

\paragraph{Contributions.} (1)~A stage-aware formalization of FM-OOD with partially observed reference distributions (Section~\ref{sec:position}). (2)~A parameter coverage ceiling result showing that model-centric methods cannot, by construction, cover inputs whose correct response requires information or computation absent from any feasible parameter configuration (Section~\ref{sec:arguments}). (3)~A precise definition of agentic OOD systems and four substantive arguments for their necessity (Section~\ref{sec:arguments}). (4)~Engagement with seven counterarguments, two of which we concede (Section~\ref{sec:counter}). (5)~Implications for evaluation, theory, and deployment, and a concrete research agenda (Section~\ref{sec:implications}).

\begin{figure}[t]
\centering
\includegraphics[width=0.8\textwidth]{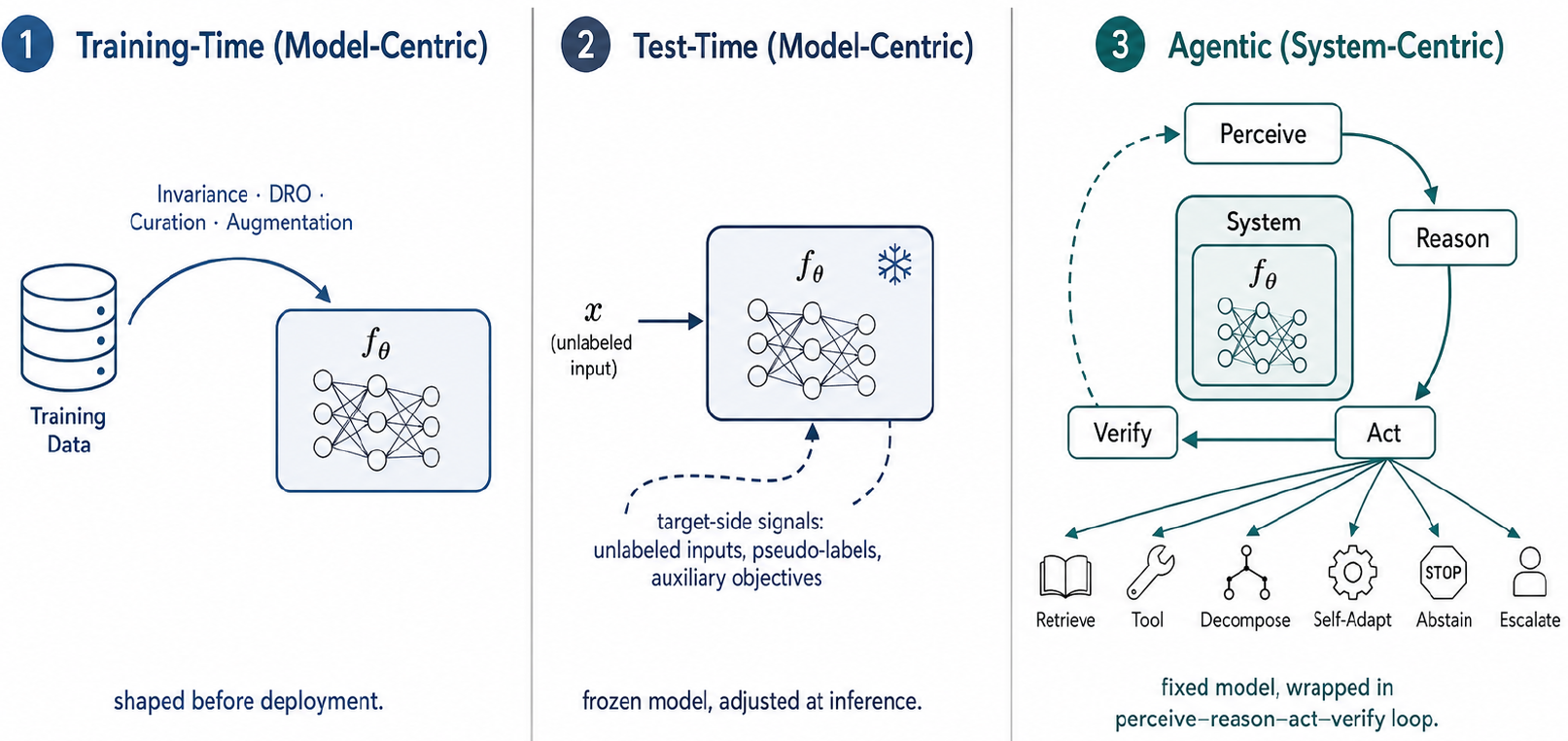}
\caption{\textbf{Three paradigms of OOD generalization for foundation models.} Training-time and test-time model-centric methods both adjust the model. Agentic methods keep the model fixed and wrap it in a perceive--reason--act--verify loop with strategies including retrieval, tools, decomposition, verification, and abstention. The paradigms overlap on inference-time model adjustment but each contains actions outside the others' reach.}
\label{fig:paradigms}
\end{figure}

\section{Position and Formal Setup}
\label{sec:position}

\subsection{Statement of Position}

We defend the following position throughout the remainder of the paper:

\begin{tcolorbox}[colback=gray!5,colframe=black!60,boxrule=0.6pt]
\textbf{Position.} OOD for foundation models is structurally distinct from classical OOD because pretraining, supervised fine-tuning, post-training, and deployment distributions are separate, interacting, and often only partially observable. In this setting, the \emph{model-centric} paradigm is insufficient: some deployment inputs require external information, computation, or deferral beyond any feasible parameter update (the \emph{parameter coverage ceiling}; Claim~\ref{claim:ceiling}). Agentic systems couple distributional self-perception with strategy selection, external action, and closed-loop verification. They are complementary to model-centric methods; neither paradigm subsumes the other.
\end{tcolorbox}

This claim is debatable: scaling may close some gaps, agentic systems may appear to rebrand TTA/RAG/tool use, and our definition may exclude many deployed systems. We address these objections in Section~\ref{sec:counter}.

\subsection{Stage-Aware Formalization of FM-OOD}
\label{sec:formalization}

The classical OOD formalization---a single training distribution $P$ and a possibly different test distribution $Q$---is too coarse for FMs because (i)~training proceeds in multiple stages with distinct distributions, (ii)~reference distributions are often only partially observed, and (iii)~deployment inputs may be OOD with respect to some stages but not others.

\paragraph{Stage-aware reference distributions.} A modern FM $f_\theta$ results from a pipeline with $K$ stages, each with its own data distribution $\mathcal{D}^{(k)}$ and support $\mathcal{S}^{(k)}$. Stages typically include $\mathcal{D}^{(\text{pre})}$ (pretraining over text, image-text pairs, or graphs); $\mathcal{D}^{(\text{sft})}$ (instruction--response pairs); $\mathcal{D}^{(\text{post})}$ (post-training, e.g., preference data for RLHF~\cite{ouyang2022training}); and optionally $\mathcal{D}^{(\text{ft})}$ (downstream fine-tuning). Deployment draws from $\mathcal{D}^{(\text{dep})}$, in general distinct from all of these.

\begin{definition}[Stage-aware OOD]
\label{def:stage-ood}
An input $x$ is \emph{out-of-distribution with respect to stage $k$}, written $x\notin\mathrm{ID}^{(k)}$, if $x$ is unlikely under $\mathcal{D}^{(k)}$ in some specified sense (e.g., $p_{\mathcal{D}^{(k)}}(x)$ below threshold $\tau_k$, or $x$ outside an empirical support estimate). $x$ is \emph{foundation-model OOD} if it is OOD with respect to some stage operationally relevant to the task.
\end{definition}

This definition has consequences the classical framing lacks. An input can be ID for one stage and OOD for another: a coding instruction in an unusual format may be ID for $\mathcal{D}^{(\text{pre})}$ but OOD for $\mathcal{D}^{(\text{sft})}$. The behavioral failure is then an instruction-following failure, not a knowledge failure, and the appropriate response differs accordingly.

\begin{definition}[Observability]
\label{def:observability}
A reference distribution $\mathcal{D}^{(k)}$ is fully observable if a representative sample $\mathcal{S}^{(k)}_{\text{obs}}$ is available; partially observable if only summary statistics or a non-representative subset are available; unobservable if no sample is available.
\end{definition}

In the FM regime, $\mathcal{D}^{(\text{pre})}$ is frequently partially observable or unobservable, especially for proprietary language and video models. Classical detectors that require training samples or known source environments therefore lose their footing. Agentic methods do not escape partial observability, but their \textsc{Perceive} stage needs only actionable signals that an input is poorly handled.

\begin{definition}[Compositional and open-ended shift]
\label{def:compositional}
A \emph{compositional shift} is a $\mathcal{D}^{(\text{dep})}$ that differs from the operationally relevant $\mathcal{D}^{(k)}$ along multiple axes simultaneously. A deployment scenario is \emph{open-ended} if the set of shifts that may occur cannot be enumerated or bounded at design time.
\end{definition}

Open-endedness sharply distinguishes FM-OOD from classical OOD. Any method whose coverage is bounded by a set fixed at design time, including the support of a training distribution, the worst-case set of a DRO ambiguity ball, or the target-domain set of an adaptation method, cannot, by construction, cover an open-ended deployment distribution.

\textbf{OOD phenomena across modalities.} Within this formalization, FM-OOD takes recognizable forms: LLMs face knowledge-boundary, temporal, task-format, and capability OOD~\cite{ji2023survey,huang2023survey,kasai2023realtime,wang2022super,dziri2023faith}; vision FMs face domain and modality shift~\cite{radford2021learning,kirillov2023segment,mahmood2021detecting,cong2022satmae,bergmann2019mvtec,zhang2024vision}; and graph FMs face compositional structural, domain, modality, and task shifts~\cite{liu2023towards,li2022outofdistribution}. The common thread is that FM-OOD is open-ended, compositional, and prone to silent failure; any OOD framework that cannot detect its own failure state is blind to the central risk of FM deployment.

\subsection{Why Existing OOD Methods Are Insufficient}
\label{sec:insufficient}

Existing OOD research is not mistaken; many of its techniques remain useful \emph{as components} within an FM system. But every strand of it encodes the same assumption: when a shift occurs, the right response is to adjust the model.

\paragraph{Training-time methods.} Invariant learning~\cite{arjovsky2019invariant,krueger2021out} requires environment labels and transferable invariances~\cite{rosenfeld2020risks}; DRO~\cite{sagawa2019distributionally,duchi2021learning} depends on a prespecified ambiguity set; augmentation~\cite{hendrycks2019augmix,zhang2017mixup} cannot add arbitrary future knowledge, capabilities, or formats; and domain generalization~\cite{zhou2022domain} assumes source domains span test-time variation.

\paragraph{Test-time methods.} TTA methods~\cite{wang2020tent,sun2020test,liang2020we}, domain adaptation~\cite{ganin2016domain,long2018conditional}, and prompt tuning~\cite{shu2022test} use target-side signals to adjust parameters, statistics, or prompts. They are effective for mild shifts but cannot supply missing knowledge, run symbolic computation, or resolve task ambiguity.

\paragraph{The common limitation.} Whether the adjusted variable is an objective, parameters, statistics, a prompt, or a feature alignment, the object of adjustment is still the model. FM-OOD needs a system-level framework for choosing \emph{what to do} under shift, including actions that do not adjust the model.

\section{Arguments: The Agentic Paradigm and Its Necessity}
\label{sec:arguments}

This section defines agentic OOD systems (Section~\ref{sec:def-agentic}), introduces the parameter coverage ceiling (Section~\ref{sec:ceiling}), clarifies the relation between paradigms (Section~\ref{sec:overlap}), and gives arguments and empirical corroboration for the agentic paradigm (Section~\ref{sec:four-args}--Section~\ref{sec:evidence}).

\subsection{What Counts as an Agentic OOD System}
\label{sec:def-agentic}

We use \emph{agentic} narrowly. Many systems use external resources without being agentic in our sense.

\begin{definition}[Agentic OOD system]
\label{def:agentic}
An OOD-handling system is \emph{agentic} if it satisfies all four:
\begin{enumerate}[leftmargin=*,topsep=2pt,itemsep=0pt]
\item \textbf{Perception.} It computes a diagnosis of whether and how the input is OOD with respect to one or more relevant $\mathcal{D}^{(k)}$, distinguishing among shift types.
\item \textbf{Strategy selection.} It has access to $\mathcal{A}$ with $|\mathcal{A}| \geq 2$ distinct response strategies and selects among them based on the diagnosis. $\mathcal{A}$ must include at least one parameter-update strategy and at least one non-parameter strategy, so selection is non-trivial.
\item \textbf{External action.} At least one strategy invokes a resource outside the model itself (retrieval, tools, decomposition, escalation).
\item \textbf{Verification with feedback.} It computes a check on its output and conditionally re-enters perception or strategy-selection on the basis of that check.
\end{enumerate}
A system performing only one of these is not agentic. A system that always retrieves before generating is \emph{retrieval-augmented} but not agentic, lacking selection and verification.
\end{definition}

This distinguishes agentic OOD from indiscriminate use of external resources: the value flows from the conjunction of perception, selection, external action, and feedback, not from any single component.

The agentic strategy space includes: \emph{Self-Adapt} (parameter or prompt update; closest to TTA~\cite{wang2020tent}), \emph{Retrieve} (knowledge access; RAG~\cite{lewis2020retrieval}), \emph{Use Tool} (external computation~\cite{schick2023toolformer}), \emph{Decompose} (sub-problems; chain-of-thought~\cite{wei2022chain}), \emph{Abstain} (selective prediction~\cite{el2010foundations}), \emph{Verify and Retry} (output checking~\cite{shinn2023reflexion}), and \emph{Escalate} (human-in-the-loop). Each addresses a distinct OOD type through a distinct mechanism; none is substitutable for the others.

\subsection{The Parameter Coverage Ceiling}
\label{sec:ceiling}

We formalize the limit on what model-centric methods can in principle achieve. Let $f_\theta:\mathcal{X}\to\mathcal{Y}$ be a FM, $L$ a task loss, $\varepsilon>0$ a tolerance.

\begin{definition}[Parameter-reachable set and coverage ceiling]
\label{def:reachable}
For a parameter neighborhood $\Theta$ of $\theta$, the \emph{$\varepsilon$-reachable set} is
$\mathcal{R}_\varepsilon(\Theta) = \{ x : \exists\, \theta' \in \Theta\ \text{s.t.}\ \mathbb{E}[L(f_{\theta'}(x), y)] \leq \varepsilon \}$.
The \emph{parameter coverage ceiling} is $\mathcal{R}_\varepsilon(\Theta_{\text{feasible}})$ where $\Theta_{\text{feasible}}$ is the set of parameter configurations reachable through any practical adjustment from $\theta$. Inputs in $\mathcal{X}\setminus\mathcal{R}_\varepsilon(\Theta_{\text{feasible}})$ are \emph{ceiling-bounded}: no model-centric method handles them at tolerance $\varepsilon$.
\end{definition}

The definition is deliberately loose about $\Theta_{\text{feasible}}$: for TTA it is reachable by a few gradient steps on target-side data; for full fine-tuning it is much larger. The key substantive claim is that even under the most generous reading of $\Theta_{\text{feasible}}$, ceiling-bounded inputs exist.

\begin{claim}[Existence of ceiling-bounded inputs]
\label{claim:ceiling}
Under standard assumptions about parameter-based representation and target-side signal availability, the following input classes are ceiling-bounded for any feasible $\Theta_{\text{feasible}}$:
(i)~inputs whose correct output requires information that is neither entailed by $\theta$ nor derivable from target-side data at deployment (e.g., facts postdating pretraining, private documents);
(ii)~inputs whose correct output requires computational precision unrealizable by any feasible $\theta'$ (e.g., exact multi-digit arithmetic, formal symbolic verification).
\end{claim}

\emph{Argument sketch.} For~(i), a parameter update can encode only information present in the data used to produce it; target-side signals at deployment (unlabeled inputs, self-supervised losses) do not carry the missing factual content, so $\mathcal{R}_\varepsilon(\Theta_{\text{feasible}})$ cannot include inputs that depend on that content. For~(ii), finite-parameter neural networks trained with standard objectives have been shown empirically to fail exact symbolic operations beyond modest scale~\cite{dziri2023faith}, and formal circuit-depth or approximation lower bounds suggest this is not merely an artifact of current training. A fully rigorous treatment would formalize ``target-side signal'' and ``feasible'' precisely; the informal claim suffices for the paradigmatic argument below.

\begin{proposition}[Coverage extension]
\label{prop:extension}
Let $\mathcal{T}$ be external resources (retrieval over corpus $\mathcal{C}$, tools $\mathcal{U}$, deferral channel $\mathcal{H}$). Define the \emph{agentic-reachable set} as $\mathcal{R}^{\text{ag}}_\varepsilon(\Theta_{\text{feasible}}, \mathcal{T}) = \mathcal{R}_\varepsilon(\Theta_{\text{feasible}}) \cup \mathcal{R}_\varepsilon^{\text{ext}}(\mathcal{T})$, where $\mathcal{R}_\varepsilon^{\text{ext}}(\mathcal{T})$ is the set of inputs for which some sequence of resource invocations in $\mathcal{T}$, possibly composed with $f_\theta$, achieves expected loss at most $\varepsilon$. Then $\mathcal{R}^{\text{ag}}_\varepsilon \supseteq \mathcal{R}_\varepsilon$, with strict inclusion whenever $\mathcal{T}$ supplies information or computation not in $\Theta_{\text{feasible}}$.
\end{proposition}

The proposition is tautological by design; we state it as such. Its content is in the design of $\mathcal{T}$. Retrieval over a daily-updated corpus provides a $\mathcal{R}_\varepsilon^{\text{ext}}$ that grows daily without retraining. Tools provide a $\mathcal{R}_\varepsilon^{\text{ext}}$ including inputs requiring exact computation. The agentic frontier sits strictly outside the parameter frontier whenever resources supply something parameters do not, which is exactly when FM-OOD is most acute.

\subsection{Relation to Model-Centric OOD: Overlap, Not Containment}
\label{sec:overlap}

\begin{claim}[Overlap, not containment]
\label{claim:overlap}
The agentic paradigm overlaps with, but does not subsume, the model-centric paradigm. The two intersect on inference-time model adjustment, which appears as the \textsc{Self-Adapt} action in agentic systems and as TTA in model-centric methods. The model-centric paradigm includes training-time interventions (invariance objectives, DRO, pretraining curation, architectural choices) not naturally expressible as inference-time agentic actions. The agentic paradigm includes inference-time actions (retrieval, tools, decomposition, verification, abstention) not expressible within model-centric methods.
\end{claim}

This corrects an earlier framing that claimed strict containment. That framing conflates \emph{when} an intervention happens (training versus inference) with \emph{what} actions are available (parameter adjustment versus external invocation). The position does not require containment. It requires only that (i) agentic methods address $\mathcal{R}_\varepsilon^{\text{ext}}(\mathcal{T}) \setminus \mathcal{R}_\varepsilon(\Theta_{\text{feasible}})$, and (ii) for inputs in the intersection, agentic methods add structural advantages (diagnostic perception, selection, verification) that model-centric methods lack. Both are true. Deployed systems will typically combine training-time robustness with inference-time agentic structure.

\subsection{Four Arguments for the Agentic Paradigm}
\label{sec:four-args}

\paragraph{Argument 1: Heterogeneous shifts require heterogeneous responses.} A user asking about yesterday's event is OOD for a structurally different reason than one requesting twenty-digit multiplication, which differs again from one submitting an unfamiliar medical image. The right responses are retrieval, calculator invocation, and cross-modal adaptation respectively. Model-centric methods apply a single mechanism to all shifts: TTA adapts when adaptation is wrong; robust training penalizes under an ambiguity set that may not match the actual shift. Condition~(2) of Definition~\ref{def:agentic} requires reasoning about shift \emph{type} and selecting accordingly.

\paragraph{Argument 2: External resources break the parameter coverage ceiling.} Proposition~\ref{prop:extension} formalizes the structural fact that agentic systems address inputs outside $\mathcal{R}_\varepsilon(\Theta_{\text{feasible}})$. For temporal knowledge, retrieval is the only solution within the formal model. For precise computation, tool use is a \emph{capability extension}, not an optimization. For graph FMs deployed in new domains, retrieving structurally analogous graphs supplies what parameters cannot produce. We do \emph{not} claim that current retrieval systems always outperform current fine-tuned models---that is an empirical question outside our scope. We claim a structural fact about coverage frontiers.

\paragraph{Argument 3: Closed-loop verification addresses silent failures.} Generative FMs that meet unknown facts produce fluent, confident, wrong answers~\cite{zhang2023siren}. Pure model-centric pipelines are \emph{open-loop}: an input enters, a forward pass runs, an output emerges, with no internal check on correctness. Even TTA's auxiliary objective is a proxy decoupled from task accuracy. Agentic systems explicitly check outputs against external evidence (self-consistency~\cite{wang2022self}, chain-of-verification~\cite{dhuliawala2023chain}, iterative reflection~\cite{shinn2023reflexion}); when checks fail, control re-enters perception. The empirical gains from these methods are largest on harder, more OOD-like inputs---the signature one would expect if they function as OOD correction mechanisms.

\paragraph{Argument 4: Compositional shifts require strategy orchestration.} Real deployments rarely present cleanly single-dimensional shifts. A clinician asking about a novel drug in a novel patient population poses simultaneous knowledge, domain, and task-format shifts. Model-centric methods, designed with one shift type in mind, degrade super-additively under composition~\cite{wiles2022fine}. Agentic systems handle compositions through \emph{decomposition} (each sub-problem is in-distribution for some appropriate strategy) and \emph{orchestration} (strategies feed each other in sequence). This sequential mixed-strategy response has no direct model-centric analogue.

\subsection{The Unified Framework}
\label{sec:framework}

\begin{figure}[t]
\centering
\includegraphics[width=0.8\textwidth]{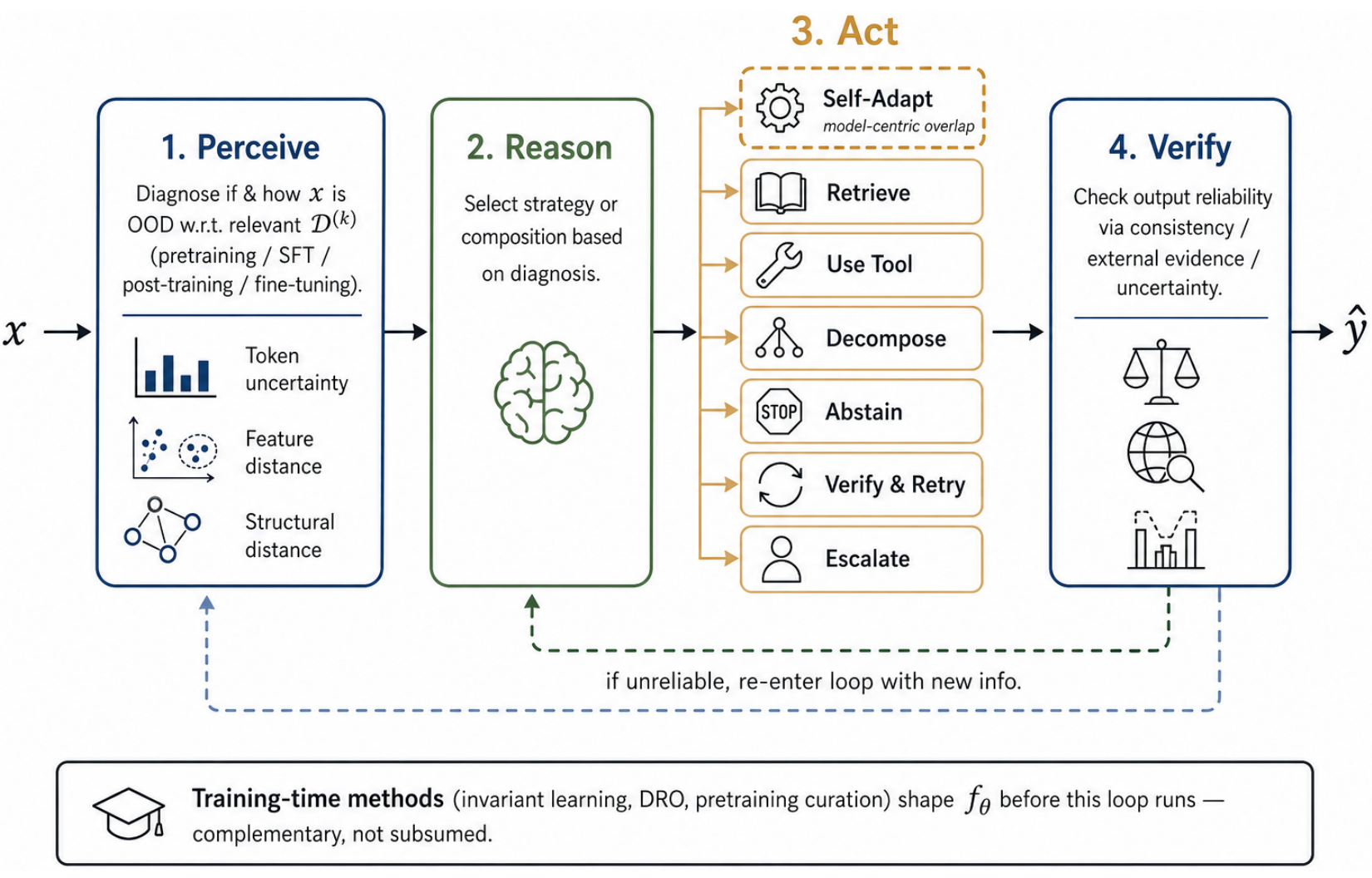}
\caption{\textbf{The unified agentic OOD framework.} An input passes through \textsc{Perceive} (diagnose OOD type w.r.t.\ relevant $\mathcal{D}^{(k)}$), \textsc{Reason} (select strategy or composition), \textsc{Act} (execute), and \textsc{Verify} (check reliability; loop back if unreliable). TTA corresponds to the \textsc{Self-Adapt} branch; training-time methods are complementary, shaping $f_\theta$ before the loop runs.}
\label{fig:framework}
\end{figure}

Synthesizing the preceding yields the framework of Figure~\ref{fig:framework}. \textsc{Perceive} diagnoses whether and how the input is OOD with respect to relevant $\mathcal{D}^{(k)}$. \textsc{Reason} selects among strategies, potentially composing several. \textsc{Act} executes the selection. \textsc{Verify} checks reliability via consistency, external evidence, or uncertainty estimates; failure triggers a return to \textsc{Reason} with new information. The framework is \emph{structurally} modality-agnostic and \emph{implementation-wise} modality-specific: a language model's perception uses token-level uncertainty; a graph model's uses structural distance from observable proxies of $\mathcal{S}^{(\text{pre})}$; a vision model's uses feature-space distance or augmentation disagreement. Strategies are also compositional: a single inference step may decompose a query, retrieve for each sub-query, use a tool on the result, and verify the composition, with each sub-strategy recursively entering its own loop.

\subsection{Empirical Corroboration: Agentic OOD Is Already Emerging}
\label{sec:evidence}

Several of the most consequential techniques in recent FM practice are \emph{partial} instances of the agentic framework of Definition~\ref{def:agentic}, even though many do not satisfy all four conditions. The convergence of partial instances on a single structural pattern is itself the evidence we report; we do not claim that any specific deployed system is fully agentic in our sense.

\emph{Retrieval-augmented generation (RAG)}~\cite{lewis2020retrieval} addresses knowledge-boundary OOD by supplying missing information from external corpora, letting models answer correctly without parameter updates. Fine-tuning to incorporate new knowledge is slow, expensive, and prone to catastrophic interference, and cannot cover information appearing after fine-tuning ends. Empirical results show that retrieval reduces hallucination on knowledge-intensive tasks~\cite{shuster2021retrieval}. Vanilla RAG is \emph{not} agentic in the full sense (no selection, no verification); the most effective recent systems augment RAG with both, moving toward fuller instantiation.

\emph{Tool-augmented FMs}~\cite{schick2023toolformer,yao2022react,qin2023toolllm} implement capability extension. The reported phenomenon that tool-augmented models outperform substantially larger untooled ones on tasks requiring specialized computation follows directly from breaking the parameter coverage ceiling of Claim~\ref{claim:ceiling}. \emph{Self-reflection methods}~\cite{wang2022self,dhuliawala2023chain,shinn2023reflexion} implement \textsc{VerifyRetry}; reported gains are largest on harder, more OOD-like inputs. \emph{Selective prediction}~\cite{kamath2020selective,varshney2023post} implements \textsc{Abstain}; in high-stakes domains, abstention is often the correct answer. \emph{Decomposition methods}~\cite{wei2022chain,zhou2022least,khot2022decomposed} address compositional OOD by reducing it to sequences of in-distribution sub-problems.

These practices converge across NLP, vision~\cite{udandarao2023sus}, robotics~\cite{kumar2021rma}, medicine~\cite{mozannar2020consistent}, and scientific discovery~\cite{boiko2023autonomous}. They were not originally designed \emph{as} OOD methods, yet all address distributional mismatch by expanding the response space beyond parameter adjustment. This is precisely when a paradigm is useful: scattered practices share a structure that should be named and developed systematically.

\section{Counterarguments and Alternative Views}
\label{sec:counter}

We engage the strongest objections to our position. We explicitly \emph{concede} Objections~2 and~4, and adjust the position accordingly.

\paragraph{Objection 1: Scaling will resolve OOD issues.} FM scaling continues to improve many OOD behaviors~\cite{kaplan2020scaling,wei2022emergent}, and we accept the premise. Three considerations bound the response. First, scaling shows diminishing returns on the OOD types of greatest practical concern: hallucination has not disappeared and, in some settings, has become more sophisticated. Second, certain OOD types are \emph{unsolvable} by model improvement alone within the formal model: temporal OOD and knowledge-boundary OOD are definitional (Claim~\ref{claim:ceiling} (i)). Third, even where scaling helps, the agentic paradigm provides an orthogonal axis. The two are complementary; progress in one does not substitute for progress in the other.

\paragraph{Objection 2 (conceded in part): Agentic approaches introduce new failure modes.} Retrieval can return wrong documents, tools can produce incorrect outputs, and multi-step errors can compound~\cite{chen2024chatgpt}. The concern is genuine; we make agentic reliability part of the research agenda (Section~\ref{sec:implications}). Three responses. First, agentic failures are \emph{observable} in a way that model-centric OOD failures typically are not: a retrieved document can be inspected, a tool output checked, a reasoning step audited, whereas a hallucinating LM offers no internal hooks. For high-stakes deployment, observable failure is preferable to silent failure. Second, the verification loop exists precisely to catch error propagation. Third, any expansion of capability brings new failure modes; the relevant comparison is failure-adjusted performance. We explicitly \emph{concede} that agentic reliability is non-trivial and must not be assumed to inherit component reliability.

\paragraph{Objection 3: This is just TTA, RAG, etc.\ rebranded.} Each strategy in our action space overlaps with an existing technique. The paradigm reduces to none of them because Definition~\ref{def:agentic} requires four elements that no individual strategy provides: stage-aware perception, strategy selection, verification with feedback, and composition. TTA does not choose \emph{when} to adapt versus retrieve. Vanilla RAG retrieves indiscriminately. The analogy is autonomous driving: it is not the union of cruise control, lane-keeping, and adaptive braking, but a system that perceives, reasons, acts across dimensions, and verifies. Agentic OOD is analogously a system that coordinates RAG, TTA, and tool use in the precise sense of Definition~\ref{def:agentic}.

\paragraph{Objection 4 (conceded): The agentic paradigm strictly contains model-centric OOD.} This was our own earlier framing, and we now regard it as incorrect. Training-time interventions---invariant learning across diverse environments, DRO, pretraining curation, architectural choices---are not naturally expressible as inference-time agentic actions; they shape $f_\theta$ \emph{before} the loop of Figure~\ref{fig:framework} runs, in ways no in-loop choice can replicate. Claim~\ref{claim:overlap} is the corrected version: the paradigms overlap on inference-time model adjustment, but each contains actions outside the other's reach. The position does not require containment; it requires only that agentic methods cover ceiling-bounded inputs, which Proposition~\ref{prop:extension} establishes whenever external resources supply what parameters do not.

\paragraph{Objection 5: A unified paradigm across modalities is vacuous.} LLM knowledge OOD, graph structural OOD, and vision domain gap differ at the surface, and a framework spanning all three may be too abstract to act on. The framework unifies at the \emph{structural} level, not the problem level. Different OOD types trigger different strategies; the framework specifies the structure within which triggering happens. Modality-specific perception and action mechanisms fill each stage. This pattern---shared structure with specialized mechanisms---is typical of successful ML paradigms (e.g., encoder--decoder, attention).

\paragraph{Objection 6: Agentic inference is too slow for real-time applications.} Full loops impose latency well beyond a single forward pass. However, the full loop need not run on every input: ID inputs pass through fast inference, and the loop is triggered only when perception flags OOD, converting overhead to a per-OOD-query cost. In domains where OOD is most prevalent and consequential (medicine, finance, law, science), correctness dominates latency in the user utility function. Adaptive-compute mechanisms~\cite{schuster2022confident} permit explicit latency--accuracy trade-offs. We concede that efficiency is an important and under-addressed research direction.

\paragraph{Objection 7: Definition~\ref{def:agentic} is restrictive enough that existing systems may not qualify, weakening the empirical evidence.} This is a fair observation. The definition is deliberately restrictive in order to distinguish agentic OOD from indiscriminate use of external resources. Many widely cited systems satisfy three of the four criteria; we flag this explicitly in Section~\ref{sec:evidence}. Two responses. First, partial instances are still informative: consistent gains from partial instances are evidence about the underlying structural pattern, even if non-exhaustive. Second, the trajectory of recent systems is precisely \emph{toward} fuller instantiation: contemporary RAG systems add selection and verification; contemporary tool-using models add perception; contemporary self-reflection systems compose with retrieval and tools. Part of our position is a \emph{prediction} that this trajectory will continue, together with an argument for driving it deliberately rather than incidentally.

\section{Implications and Research Agenda}
\label{sec:implications}

Recognizing agentic OOD as a first-class paradigm reshapes research priorities in six concrete directions and carries concrete implications for evaluation, theory, and deployment.

\paragraph{(1) OOD-aware perception.} Mechanisms that distinguish knowledge-boundary OOD (w.r.t.\ $\mathcal{D}^{(\text{pre})}$) from task-format OOD (w.r.t.\ $\mathcal{D}^{(\text{sft})}$), from capability-limit OOD, from compositional shift. This is sharply harder when reference distributions are only partially observable. Candidate approaches: contrastive probing, auxiliary heads predicting shift type, reasoning traces as diagnostic signals, and uncertainty decomposition into aleatoric vs.\ distributional components.

\paragraph{(2) Strategy selection and orchestration.} Meta-learning over strategy choice framed as a discrete decision problem; reinforcement learning with correctness--efficiency rewards; cost--benefit analyses trading retrieval latency against error reduction. The conceptual challenge is that strategies \emph{interact}: a principled theory of composition is required and does not yet exist.

\paragraph{(3) Agentic reliability under OOD.} Verification mechanisms with formal guarantees; error-propagation analysis across multi-step pipelines; robustness to adversarial inputs targeting perception or selection (e.g., prompt injection into retrieved documents). This directly addresses Objection~2.

\paragraph{(4) Evaluation.} Benchmarks that explicitly separate model-centric-solvable from agentic-required OOD; stage-aware benchmarks labeling each input by the violated reference distribution; metrics for verification faithfulness, selection appropriateness, and compositional coverage---rather than end-to-end accuracy alone. Current OOD benchmarks~\cite{koh2021wilds,hendrycks2019benchmarking} largely conflate these categories.

\paragraph{(5) Efficiency.} Adaptive compute, early exit within the loop~\cite{schuster2022confident}, threshold calibration for invoking the agentic loop, and caching and amortization of repeated sub-strategies. This directly addresses Objection~6.

\paragraph{(6) Theory.} Generalization bounds for systems that invoke external resources and verify outputs; conditions under which agentic adaptation provably outperforms any static model; sample-complexity implications of retrieval access. Proposition~\ref{prop:extension} is only a starting point; a full theory would quantify $\mathcal{R}^{\text{ext}}_\varepsilon(\mathcal{T})$ as a function of $\mathcal{T}$, characterize when verification provides accuracy guarantees, and bound compositional error propagation.

\paragraph{Implications for deployment.} For practitioners, the position implies that FM deployments in open-world settings should be designed as agentic systems \emph{from the outset}, rather than as monolithic models with retrieval or tool use bolted on. For evaluation, model-level benchmarks are insufficient to certify FM-OOD performance: system-level benchmarks that include perception, selection, and verification are required. For theory, generalization analysis must extend beyond parameter-based hypothesis classes to include external-resource-augmented classes.

\section{Conclusion}
\label{sec:conclusion}

OOD for foundation models is structurally distinct from classical OOD: stage-aware (multiple interacting reference distributions), often partially observable, open-ended, compositional, and prone to silent failure. Model-centric methods---whether training-time invariance, robust optimization, pretraining curation, or test-time adaptation---share an assumption that fails in this regime: that adjusting the model is always the right response. We have argued that agentic systems, \emph{narrowly defined} by perception, strategy selection, external action, and verification with feedback, form the missing paradigm. The two paradigms overlap on inference-time model adjustment but neither subsumes the other; agentic methods cover the gap between the parameter coverage ceiling and the agentic-reachable set, a gap that model-centric methods cannot close by construction. The paradigm is already implicit in much successful FM practice but has not been recognized as an OOD framework, and its development has been scattered across communities. We urge the community to explicitly recognize foundation-model OOD as a distinct problem, unify its scattered agentic instantiations, and develop agentic OOD as a first-class research direction alongside the model-centric approaches that currently dominate the literature.



\bibliographystyle{IEEEtran}
{
\small
\bibliography{ref}
}


\end{document}